\documentclass[letterpaper]{article} 
\usepackage{aaai2026}  
\usepackage{times}  
\usepackage{helvet}  
\usepackage{courier}  
\usepackage[hyphens]{url}  
\usepackage{graphicx} 
\usepackage{natbib}  
\usepackage{caption} 
\usepackage{algorithm}
\usepackage{algorithmic}
\usepackage{algorithmic}

\usepackage{amsmath}
\usepackage{dsfont}
\usepackage{multirow} 
\usepackage{amssymb}

\usepackage{newfloat}
\usepackage{listings}
\DeclareCaptionStyle{ruled}{labelfont=normalfont,labelsep=colon,strut=off} 
\floatstyle{ruled}
\newfloat{listing}{tb}{lst}{}
\floatname{listing}{Listing}
\title{RaLiFlow: Scene Flow Estimation with 4D Radar and LiDAR Point Clouds}
\author{
    Jingyun Fu\textsuperscript{\rm1},
    Zhiyu Xiang\textsuperscript{\rm 1}\thanks{Corresponding authors.},
    Na Zhao\textsuperscript{\rm 2}\footnotemark[\value{footnote}]
}
\affiliations{
    \textsuperscript{\rm 1}Zhejiang University\\
    \textsuperscript{\rm 2}Singapore University of Technology and Design\\
    fujingyun@zju.edu.cn, xiangzy@zju.edu.cn, na\_zhao@sutd.edu.sg
}

\begin{document}

\maketitle

\begin{abstract}
Recent multimodal fusion methods, integrating images with LiDAR point clouds, have shown promise in scene flow estimation. However, the fusion of 4D millimeter wave radar and LiDAR remains unexplored. Unlike LiDAR, radar is cheaper, more robust in various weather conditions and can detect point-wise velocity, making it a valuable complement to LiDAR. However, radar inputs pose challenges due to noise, low resolution, and sparsity. 
Moreover, there is currently no dataset that combines LiDAR and radar data specifically for scene flow estimation. 
To address this gap, we construct a Radar-LiDAR scene flow dataset based on a public real-world automotive dataset. We propose an effective preprocessing strategy for radar denoising and scene flow label generation, deriving more reliable flow ground truth for radar points out of the object boundaries. Additionally, we introduce RaLiFlow, the first joint scene flow learning framework for 4D radar and LiDAR, which achieves effective radar-LiDAR fusion through a novel Dynamic-aware Bidirectional Cross-modal Fusion (DBCF) module and a carefully designed set of loss functions. The DBCF module integrates dynamic cues from radar into the local cross-attention mechanism, enabling the propagation of contextual information across modalities. Meanwhile, the proposed loss functions mitigate the adverse effects of unreliable radar data during training and enhance the instance-level consistency in scene flow predictions from both modalities, particularly for dynamic foreground areas.
Extensive experiments on the repurposed scene flow dataset demonstrate that our method outperforms existing LiDAR-based and radar-based single-modal methods by a significant margin.
\end{abstract}

\begin{links}
    \link{Code}{https://github.com/FuJingyun/RaLiFlow}
\end{links}

\section{Introduction}
Scene flow estimation extracts 3D motion fields of the dynamic environment from consecutive input frames, ranging from monocular images~\cite{liang2025zero}, stereo pairs~\cite{jiao2021effiscene}, LiDAR point clouds~\cite{kim2025flow4d} and emerging 4D radar data~\cite{ding2022self}. It provides class-agnostic information for scene understanding~\cite{zhao2021few, li2024end, han2024dual, wang2025augrefer} and supports higher-level tasks such as 3D detection~\cite{sheng2022rethinking, erccelik20223d, sheng2025ct3d++}, 3D panoptic segmentation~\cite{pan2025images}, multi-object tracking~\cite{najibi2022motion}, autonomous navigation~\cite{chen2025ssf} and 4D occupancy forecasting~\cite{guo2024fsf}. Recently, LiDAR-based scene flow estimation~\cite{lin2024flowmamba, luo2025mambaflow} has become mainstream, with studies~\cite{liu2022camliflow, wan2023rpeflow} integrating complementary information from 2D images or event cameras to enhance scene flow learning. However, since 2D data are not naturally aligned with 3D point clouds, these methods rely on 3D projection-based correspondences, leading to potential information loss during cross-modal fusion. Moreover, cameras and LiDAR sensors are susceptible to poor environmental conditions like storms and fog. In contrast, 4D millimeter wave radar, which shares the same point cloud format as LiDAR, is more robust to adverse weather due to its strong penetrative capabilities. Additionally, 4D radar is cost-effective and provides radial velocity measurement, offering direct motion cues for scene flow prediction. As such, 4D radar is a valuable complement to LiDAR for scene flow learning.

\begin{figure}
    \centering
    \includegraphics[width=0.99\linewidth]
    {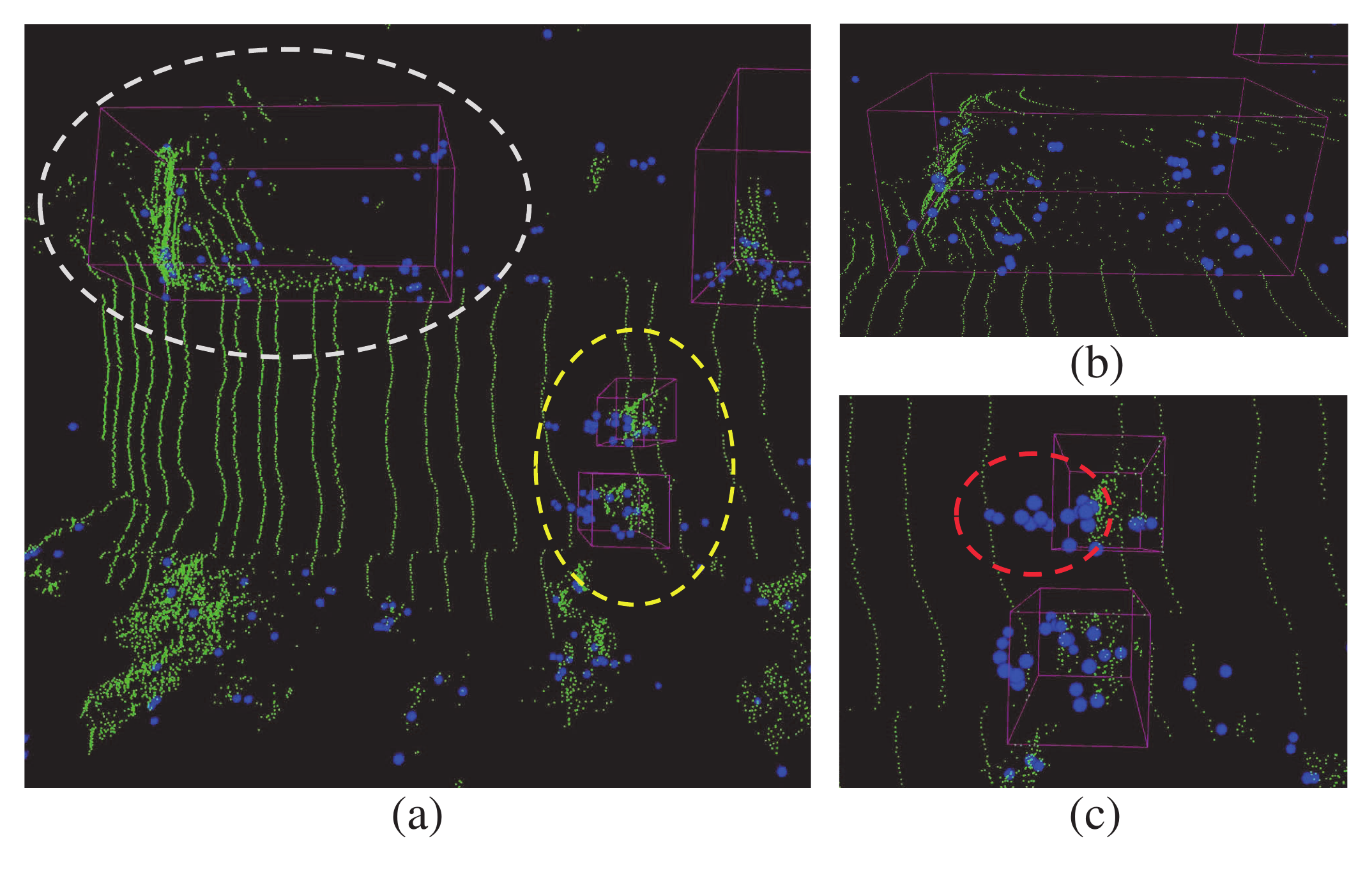}
    \caption{(a) Visualized comparison between 4D radar (blue) and LiDAR (green) point clouds. The magenta boxes are the tracked 3D bounding boxes from VoD dataset~\cite{palffy2022multi} ground truth. (b) and (c) show the zoomed-in display results of the vehicle area (circled in white) and the pedestrian area (circled in yellow), respectively. It can be seen that the radar point cloud is much sparser and more noisy than LiDAR point cloud, with many radar points on the target object falling out of the bounding boxes (circled in red).}
    \label{fig:rawdata}
\end{figure}

Despite the complementary nature of radar and LiDAR, their fusion remains unexplored for scene flow estimation due to two main challenges.
First, radar-based scene flow learning is intractable due to the sparsity, inherent noise, and low resolution of the radar data (Fig.~\ref{fig:rawdata}). Up to now, very few scene flow estimation methods \cite{ding2022self, ding2023hidden, ding2024milliflow} have involved radar data, and its potential remains underutilized. Second, presently there is no real-world scene flow dataset that contains both radar and LiDAR data. Since manual scene flow labeling is quite expensive, some previous studies \cite{jund2021scalable, ding2022self, ding2023hidden,li2023fast,zhang2024deflow} generate 
scene flow labels based on synchronized odometry data and rigid transformations between tracking boxes across consecutive frames. Although usable LiDAR scene flow labels can be produced from box-to-box correspondences, this processing method is suboptimal for radar because many dynamic radar points are outside the box, as shown in Fig.~\ref{fig:label}. These radar points will be incorrectly classified as static, resulting in unreliable flow labels and further negatively impacting scene flow learning in both training and evaluation stages.

To address the dataset gap, we derive a Radar-LiDAR scene flow dataset from VoD dataset~\cite{palffy2022multi} with our effective preprocessing strategy. We propose a cross-modal hybrid radar denoising method that first identifies radar outliers by examining their absolute radial velocity (ARV). 
After that, the contextual information from the aligned LiDAR point cloud is used as a soft threshold to filter out isolated and irregular radar points. In addition, we propose a two-pronged strategy to generate more reliable radar scene flow labels. Specifically, we first calculate the rigid flow vectors for in-box points. Then, we identify the misclassified dynamic radar points based on the 3D spatial distance from each point to its nearest bounding box, the category label of the tracking box, and the consistency between ARV and the radial projection of the expected scene flow.

With our created dataset, it becomes possible to explore the radar-LiDAR fusion strategy for scene flow learning.
Note that existing radar-LiDAR fusion methods are primarily designed for 3D object detection \cite{wang2022interfusion, li2023pillardan, tai2024fusing, xu2024sckd,lin2024rcbevdet} and cannot be directly applied to scene flow estimation. These methods focus on object-level feature extraction, while scene flow estimation requires fine-grained, class-agnostic dynamic details at the point level. Furthermore, the global attention mechanism~\cite{liu2021global} commonly used in these methods tends to prioritize target object regions and often discard irrelevant scene details.
This information loss can lead to significant performance degradation for scene flow prediction.

To address the method gap, we propose RaLiFlow, the first radar-LiDAR fusion framework for scene flow estimation, leveraging the strengths of both sensors. Pillar-based 2D features are extracted for 4D millimeter wave radar and LiDAR inputs, and complementary information interaction between these two modalities is enhanced with a novel Dynamic-aware Bidirectional Cross-modal Fusion (DBCF) module. The DBCF module performs local cross-attention on the 2D feature maps of the two modalities under the guidance of radar dynamics. In particular, the velocity measurement inherent in radar data can initially indicate the dynamic foreground areas in the scene, which is most important and challenging for scene flow estimation. Inspired by this, the DBCF module computes a Gaussian heatmap based on the distance from each 2D grid center to its nearest dynamic radar grid, which provides attentive weight in the subsequent cross-attention operation. Moreover, we develop a set of loss functions to discard unreliable radar points in loss calculation and enforce scene flow coherence within dynamic foreground instances for both modalities.

Our main contributions are summarized as follows:
\begin{itemize}
\item 
We create a Radar-LiDAR scene flow dataset from VoD dataset with our effective preprocessing strategy, which provides more reliable radar scene flow labels.
\item 
We present RaLiFlow, the first radar-LiDAR fusion network for scene flow estimation, exploiting the complementary information between the two modalities.
\item
We specially design a novel Dynamic-aware Bidirectional Cross-modal Fusion (DBCF) module which integrates radar dynamics into local cross-attention mechanism to enhance cross-modal information interaction. 
\item
We develop effective loss functions to filter out unreliable radar supervision signals for training and encourage instance-level flow consistency of the two modalities. 
\item 
Extensive experiments on the repurposed scene flow dataset demonstrate that our method achieves state-of-the-art performance, surpassing existing single-modal methods by significant margins.
\end{itemize}

\section{Related work}
\label{sc:related}
\paragraph{LiDAR Scene Flow Estimation.}
Nowadays, learning-based methods have been dominant for LiDAR scence flow estimation~\cite{liu2019flownet3d, wu2020pointpwc,cheng2022bi,cheng2023multi,wang2023ihnet,lin2024flowmamba} on synthetic datasets. 
However, these methods only support limited size of input data and suffer from the domain gap between synthetic and real data. To handle large-scale real-world LiDAR point clouds, a series of pillar-based~\cite{jund2021scalable,zhang2024deflow,zhang2024seflow,khoche2025ssf} and voxel-based~\cite{li2022sctn,kim2025flow4d,luo2025mambaflow} pipelines rasterize world space into BEV grid and sparse volumetric grid, efficiently abstracting 2D and 3D grid-level features for the point clouds. Leveraging runtime optimization, another line of work~\cite{li2021neural,li2023fast,vedder2023zeroflow,vedder2024scene} is applicable to large-scale point cloud input but requires substantial computational resources. Moreover, several recent works~\cite{liu2022camliflow,wan2023rpeflow} explore multimodal fusion for scene flow estimation and incorporate 2D modalities like images or event cameras with 3D LiDAR point clouds. However, how to fuse 4D millimeter wave radar and LiDAR data for scene flow estimation remains unknown.

\paragraph{Radar Scene Flow Estimation.}
Despite remarkable progress in LiDAR-based scene flow estimation, radar-based scene flow estimation remains largely underexplored. RaFlow~\cite{ding2022self} first introduces a self-supervised framework for scene flow prediction on radar point clouds, but suffers from the absence of effective supervision signals. MilliFlow~\cite{ding2024milliflow} focuses on human motion estimation, relying on 3D human skeleton labels from RGB-D images for radar scene flow prediction.
CMFlow~\cite{ding2023hidden} proposes a multitask architecture for radar scene flow learning, obtaining cross-modal supervision from ego motion, dynamic segmentation, optical flow and 3D scene flow labels generated by existing optical flow estimation and 3D tracking networks. Therefore, CMFlow is affected by the performance of these related models. Moreover, these existing radar-based methods can only handle a small number of input radar points. In contrast, our RaLiFlow can dynamically adapt to large scale real-world inputs and does not rely on existing networks for label generation.

\paragraph{3D Scene Flow Annotation.}
As annotating dense scene flow on real-world data is extremely expensive, synthetic datasets are widely used in early scene flow estimation methods~\cite{geiger2013vision,mayer2016large,jin2022deformation}. However, the distribution pattern of virtual point clouds differs from real data, and the commonly created point-level one-to-one correspondence between virtual frames is unrealistic. To generate flow ground truth from real-world datasets, previous  methods~\cite{jund2021scalable,ding2022fh,zhang2024deflow,zhang2024seflow,khoche2025ssf,li2023fast} compute rigid scene flow from tracking boxes for in-box points and assign ego-motion vectors to points outside the boxes based on synchronized odometry. While this approach generates usable flow labels for LiDAR data, it is problematic for radar as many dynamic radar points may fall outside of the tracking boxes. As of now, though recent works~\cite{lin2024icp,jiang20243dsflabelling} have developed automatic flow annotation pipelines for LiDAR datasets, obtaining reliable scene flow ground truth for radar remains challenging.

\paragraph{Radar-LiDAR Fusion for 3D Detection.}
Recently, radar-LiDAR fusion~\cite{wang2022interfusion,wang2023bi,li2023pillardan,tai2024fusing,lin2024rcbevdet} has proven effective in 3D object detection. However, these detection methods focus on object-level scene representation and the commonly used global attention may damage the point-level details in the scene, thus these methods cannot be simply applied to fine-grained scene flow estimation scenario. Currently, the dynamic cues inherent in radar have not been effectively exploited in cross-modal feature fusion.

\section{Methodology}
\label{sc:approach}
Given two consecutive frames of 4D radar and LiDAR data, let the radar point clouds in the source and target frames be denoted as $\mathbf{P}^{R}$ and $\mathbf{Q}^{R}$, respectively, and the corresponding LiDAR point clouds as $\mathbf{P}^{L}$ and $\mathbf{Q}^{L}$. The joint scene flow learning for both modalities aims to predict the point-wise scene flow vector $\mathbf{v}_{i} \in \mathbf{V}^{R}$ for each radar point $\mathbf{p}_{i} \in \mathbf{P}^{R}$, and $\mathbf{v}_{j} \in \mathbf{V}^{L}$ for each LiDAR point $\mathbf{p}_{j} \in \mathbf{P}^{L}$ in the source frame. The scene flow vectors can be decomposed into ego vehicle's motion and the absolute motion of each point:
\begin{equation}
    \mathbf{V}^{R,L} = \mathbf{V}^{R,L}_{ego} +  \hat{\mathbf{V}}^{R,L}.
\end{equation}
For a fair comparison with previous studies~\cite{jund2021scalable,zhang2024deflow,li2023fast}, we assume that the vehicle's ego motion is given and the output of our network is the relative scene flow $\hat{\mathbf{V}}^{R,L}$, which is later integrated with the known ego-motion flow to form the final scene flow prediction $\mathbf{V}=\{\mathbf{V}^{R},\mathbf{V}^{L}\}$. 

\noindent The goal of our task is to minimize the total End Point Error (EPE) between $\mathbf{V}$ and the absolute ground truth flow $\mathbf{V}^{gt}$: 
\begin{align}
\nonumber
    \min (\underbrace{ \frac{1}{\left\|\mathbf{V}^{R}\right\|} \sum_{\mathbf{v}_{i} \in \mathbf{V}^{R}} \left\| \mathbf{v}_{i} - \mathbf{v}_{i}^{gt} \right\|_2 }_{\text{Radar EPE}}+\\
    \underbrace{ \frac{1}{\left\|\mathbf{V}^{L}\right\|} \sum_{\mathbf{v}_{j} \in \mathbf{V}^{L}} \left\| \mathbf{v}_{j} - \mathbf{v}_{j}^{gt} \right\|_2 }_{\text{LiDAR EPE}}).
\end{align}

\subsection{Radar-LiDAR-based Scene Flow Dataset}
\label{sc:data} 
This section introduces how to create a Radar-LiDAR-based scene flow dataset from the real-world automotive VoD dataset~\cite{palffy2022multi} with our effective data preprocessing strategy.

\paragraph{Ground Point Removal.}
Due to the aperture problem, estimating the local motion of points on flat ground surfaces is challenging. Therefore, ground removal is crucial for accurate scene flow estimation~\cite{najibi2022motion, lin2024icp, zhang2024deflow, zhang2024seflow, khoche2025ssf}. In our approach, we first project the radar point clouds into the LiDAR coordinate system using the given sensor extrinsics. Then, we apply the method from~\cite{himmelsbach2010fast} to remove the ground points from the combined radar and LiDAR scans.

\begin{figure}
    \centering
    \includegraphics[width=0.99\linewidth]
    {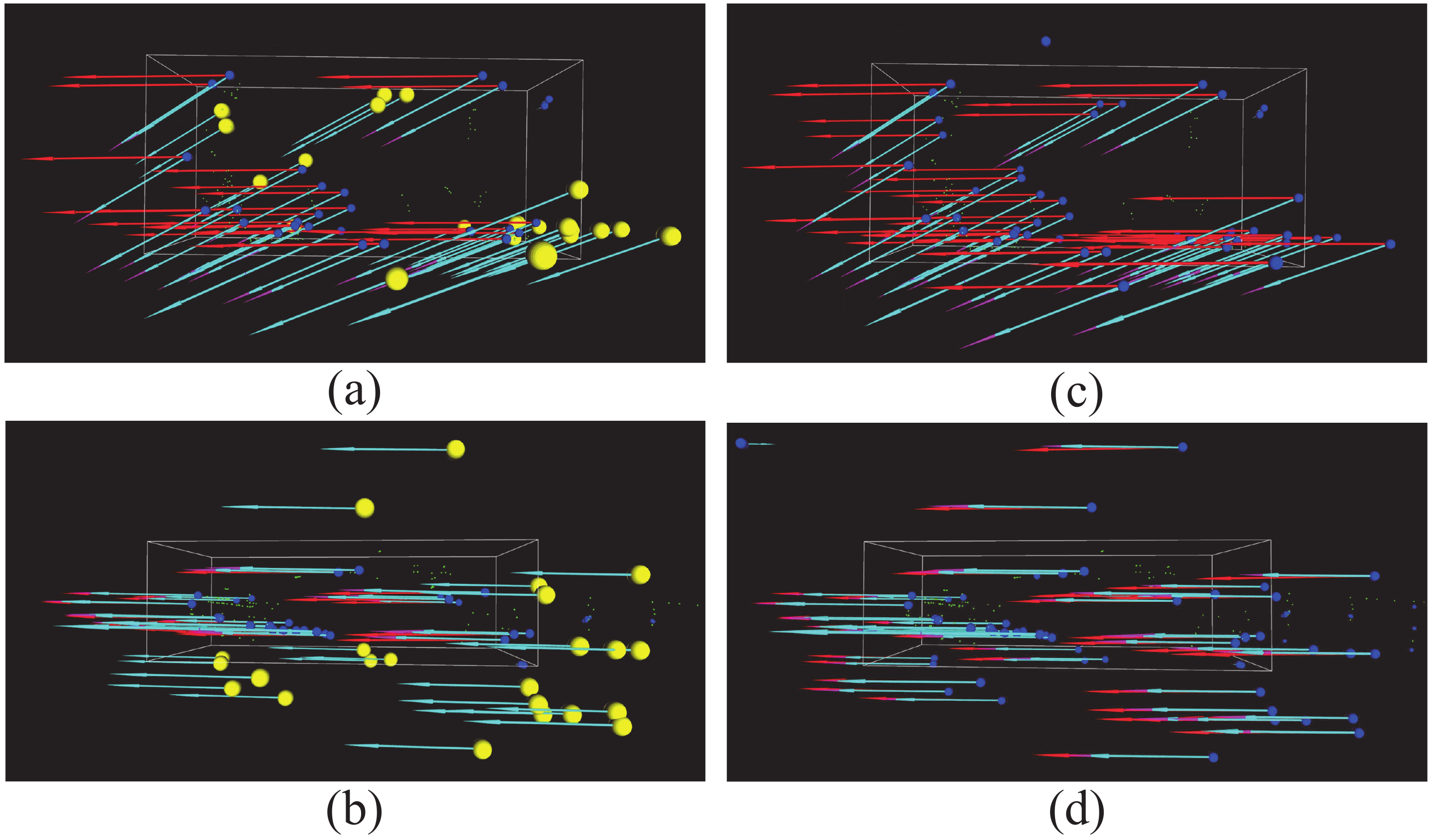}
    \caption{Comparison between the radar scene flow labels (red arrow) generated by our method (c and d) and previous methods (a and b). The illustration shows a moving car (white bounding box) from both bird's-eye (a and c) and side views (b and d), with blue radar points and green LiDAR points correspond to it. The magenta arrows represent the radial projection of the generated radar scene flow labels, and cyan arrows represent point-wise ARV captured by the radar sensor. The highlighted yellow radar points indicate significant differences between the radial projection of abosolute scene flow and ARV.}
    \label{fig:label}
\end{figure}

\begin{figure*}[t]
\centering
    \includegraphics[width=0.99\linewidth]
    {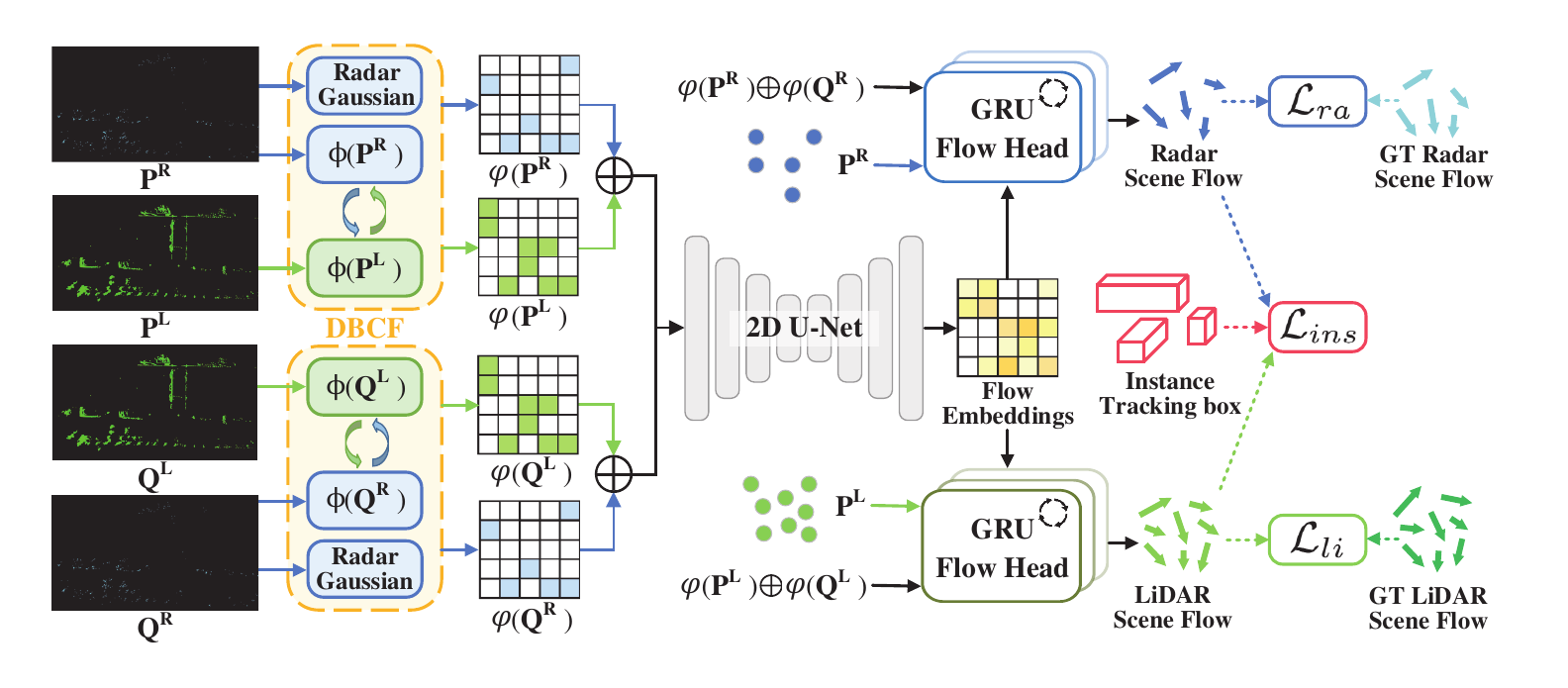}
    \caption{The overall architecture of RaLiFlow network. $\phi(\cdot)$ denotes pillar-based 2D feature map extracted for each input point cloud, and $\varphi(\cdot)$ represents the fused 2D feature map produced by DBCF module. GT stands for ground truth and $\oplus$ represents concatenation.}
    \label{fig:RLFlow}
\end{figure*}

\paragraph{Cross-Modal Hybrid Radar Denoising.}
After ground removal, we filter out radar outliers using a combination of a hard threshold based on the radar's inherent motion information and spatial context (soft threshold) provided by the LiDAR data. The VoD dataset~\cite{palffy2022multi} provides both relative and absolute radial velocity (ARV) measurements for each radar point. We establish a hard ARV threshold $\mu$ based on the average ARV of the dynamic radar points in each frame to filter out 1-10 abnormal radar points:
\begin{equation}
\mu=ARV_{avg}+\vartheta_{thre}.
\end{equation}
where $\vartheta_{thre}$ is set to $10\mathrm{~m/s}$ for computation efficiency and radar points with $ARV>0.5\mathrm{~m/s}$ are identified as dynamic.

Additionally, we leverage contextual information from the aligned LiDAR data to remove isolated radar points, since
radar point clouds are much sparser (500+ vs. 9000+ in our case) and noisier than LiDAR under normal conditions. Specifically, we project the combined point clouds onto a bird’s-eye view (BEV) grid with a grid size of $0.8\mathrm{~m}$. If no LiDAR points are found in a neighboring $3 \times 3$ grid area around a given radar point, that radar point is discarded. This loose search range ensures that only a few irregular radar points are removed and valid radar points are reserved even when the radar penetrates obstacles.

\paragraph{Scene Flow Label Generation.}
\label{sf_gene}
We generate in-box scene flow labels for both LiDAR and radar point clouds based on the rigid transformation between the corresponding tracking boxes in the frame sequence. In this process, points outside the labeled boxes are assigned flow values of $0\mathrm{~m/s}$.
To address the noisy nature of radar points that may fall outside the boxes, we apply a two-pronged strategy to re-estimate the flow for those radar points outside the labeled boxes but with ARV\textgreater$0.5\mathrm{~m/s}$. First, we calculate the distance of each candidate radar point to the center of its nearest bounding box. Given that tracking boxes have different sizes across categories, we use the class labels in the VoD dataset to set category-specific distance thresholds based on statistical results from the training set. This allows us to determine whether a radar point is a valid dynamic point by considering its distance to the nearest bounding box and the class of the box. Additionally, we check the consistency between the ARV and the candidate radial flow for these radar points, with a conservative threshold of $\gamma_{thre}=1\mathrm{~m/s}$:
\begin{equation}
\label{eq:consistency}
\lvert radial(\mathbf{v}_{i})-\mathrm{ARV}_{i} \rvert < \gamma_{thre}.
\end{equation}
Candidate radar points that do not satisfy this condition are classified as static background points. As illustrated in Fig.~\ref{fig:label}, our two-pronged approach generates more reliable scene flow ground truth for radar inputs, mitigating the negative effects of poor radar label quality in both the training and evaluation phases.

\subsection{RaLiFlow Network}
\paragraph{Framework Architecture.}
Fig.~\ref{fig:RLFlow} illustrates the overall framework of our proposed RaLiFlow. 
Given two consecutive point cloud frames of radar ($\mathbf{P}^{R}$, $\mathbf{Q}^{R}$) and LiDAR data ($\mathbf{P}^{L}$, $\mathbf{Q}^{L}$), we first extract a pseudo 2D feature map for each point cloud using pillar-based encoder~\cite{desai2020review}, denoted as $\phi(\cdot)$. 
A Gaussian heatmap generated from radar dynamics and a pair of pseudo 2D features from different modalities in each frame are then fed into our Dynamic-aware Bidirectional Cross-modal Fusion (DBCF) module.
The DBCF module integrates radar dynamics into bidirectional local cross-attention between radar and LiDAR features to generate fused feature maps for each modality, denoted as $\varphi(\cdot)$.
Here, $\varphi(\cdot)$ represents the combined operator of $\phi(\cdot)$ and the DBCF module. Subsequently, these fused features are then concatenated and fed into a U-Net backbone to produce a 2D flow embedding map.
Finally, the flow embeddings, raw source point clouds, and concatenated fused feature maps from two frames are fed into two GRU-based flow heads~\cite{cho2014properties,zhang2024deflow}, which iteratively generate the final scene flow predictions for each modality separately. Our RaLiFlow is trained using three loss functions: the LiDAR flow loss $\mathcal{L}_{li}$, the masked radar flow loss $\mathcal{L}_{ra}$, and the instance-level flow consistency loss $\mathcal{L}_{ins}$, as detailed in Sec.~\ref{sec:loss}.

\paragraph{Dynamic-aware Bidirectional Cross-modal Fusion.}
To promote the interaction of complementary information across modalities, cross-attention provides a straightforward approach for fusing radar and LiDAR data. However, to better leverage the inherent motion information in 4D millimeter-wave radar and mitigate detail loss caused by the large receptive fields in global attention-based operations, we propose a novel Dynamic-aware Bidirectional Cross-modal Fusion (DBCF) module. This module integrates radar dynamics into a localized cross-attention mechanism. Specifically, radar dynamics-aware local cross-attention is applied bidirectionally: radar serves as the query to fuse local neighbors in the LiDAR domain, and vice versa, LiDAR serves as the query to fuse local neighbors in the radar domain.

\begin{figure}[t]
    \centering
    \includegraphics[width=0.99\linewidth]{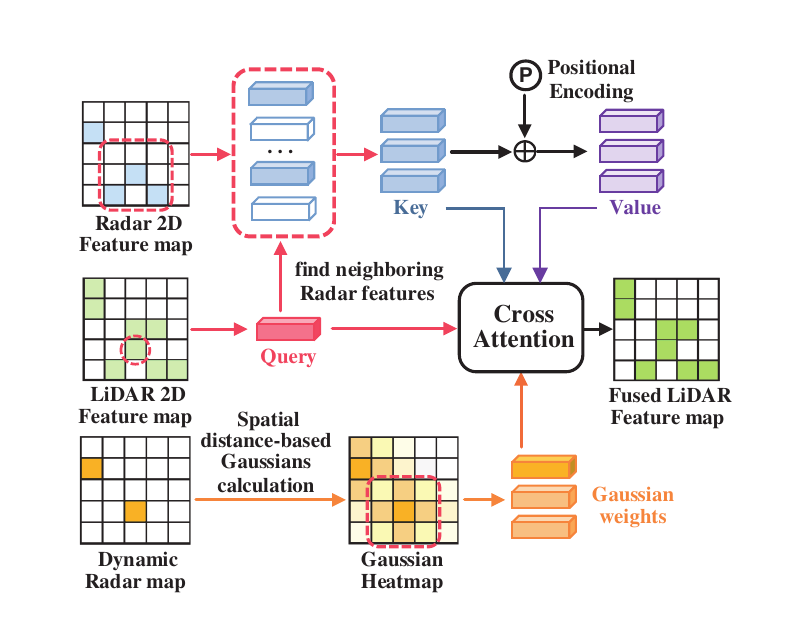}
    \caption{
    The process of one-directional dynamic-aware cross-modal fusion, with LiDAR feature serving as the query.}
    \label{fig:DBCF}
\end{figure}

Fig.~\ref{fig:DBCF} illustrates the detailed process of how to conduct the one-directional radar-to-LiDAR fusion.
In this case, each non-sparse feature in the LiDAR feature map will be selected as query $Q$, and the non-sparse features in the $3\times3$ neighboring area of the radar feature map will be collected as keys $K$. After that, a relative 2D position encoding will be added to these neighboring features, yileding values $V$ for the following cross-attention operation. 

The absolute radial velocity (ARV) measurement from 4D radar naturally facilitates dynamic perception of the scene, allowing us to focus on challenging areas that are more likely to be dynamic when performing cross-attention operations. As shown in Fig.~\ref{fig:DBCF}, the dynamic radar map represents pillars containing radar points with ARV\textgreater$0.1\mathrm{~m/s}$. We calculate the 2D distance $D$ from each grid center to its nearest dynamic radar grid center, and the Gaussian heatmap $G$ is then generated as follows: 
\begin{equation}
    G = \exp(-D^2/\sigma^2),
\end{equation}
where $\sigma$ determines the bandwidth of the Gaussian distribution. 
The generated $G$ assigns higher weight to 2D features locate near the radar dynamic area when integrating modality $\mathcal{x}$ into modality $\mathcal{y}$:
\begin{equation}
\mathcal{A}_{\mathcal{x}\rightarrow \mathcal{y}}(Q^{\mathcal{y}},K^{\mathcal{x}},V^{\mathcal{x}}|G) = \mathrm{Softmax}( \frac{Q^{\mathcal{y}}(K^{\mathcal{x}})^{T}}{\sqrt{c}}\odot{G})V^{\mathcal{x}},
\end{equation}
where $c$ is the channel dimension and $\odot$ is the element-wise multiplication. 
Then the fused 2D feature maps $\varphi(\cdot)$ in each frame can be calculated from the previous pseudo 2D feature maps $\phi(\cdot)$. For the source frame, $\varphi(\mathbf{P}^R)$ and $\varphi(\mathbf{P}^L)$ are constructed as follows:
\begin{equation}
    \varphi(\mathbf{P}^R) = \mathcal{A}_{L\rightarrow R}(Q^{R},K^{L},V^{L}|G)+\phi(\mathbf{P}^{R}),
    \label{att_lr}
\end{equation}
\begin{equation}
    \varphi(\mathbf{P}^L) = \mathcal{A}_{R\rightarrow L}(Q^{L},K^{R},V^{R}|G)+\phi(\mathbf{P}^{L}).
    \label{att_rl}
\end{equation}

\subsection{Loss Function}
\label{sec:loss}

\paragraph{LiDAR Flow Loss.}
\label{pa:li_loss}
As most of the LiDAR points in real scenarios are static, we adopt the loss in ~\cite{zhang2024deflow} to balance the contribution of the input points. Since point-wise speed $\textbf{s}(p_j)$ can be directly obtained from current flow prediction, the points in $\mathbf{P}^{L}$ are divided into three point sets with different ranges of motion speeds:
\begin{equation}
    p_{j}\in \begin{cases} \mathbf{S}_1^L &  \text { if } \textbf{s}(p_j)>1.0 \mathrm{~m} / \mathrm{s} \\ 
     \mathbf{S}_2^L &  \text { if } \textbf{s}(p_j)<0.4 \mathrm{~m} / \mathrm{s} \\ 
    \mathbf{S}_3^L &  \text { o.w. }\end{cases}.
    \label{sp:speed_pt}
\end{equation}
where $\mathbf{S}_1^L \cup \mathbf{S}_2^L \cup \mathbf{S}_3^L = \mathbf{P}^L$. 
The LiDAR flow loss is calculated as:
\begin{equation}
    \mathcal{L}_{li} = \sum_{k=1}^{3}\frac{1}{|\mathbf{S}^{L}_{k}|} \sum_{p_{j} \in \mathbf{S}^{L}_{k}} \left\| \mathbf{V}^{L}(p_{j})- \mathbf{V}^{L}_{gt}(p_{j})\right\|_2.
\label{eq:li_loss}
\end{equation}

\paragraph{Masked Radar Flow Loss.}
Since radar point clouds are much sparser and noisier, we filter out unreliable radar points with mismatched flow labels and ARV during loss calculation to reduce its negative impact on network training. Using Eq.~\eqref{eq:consistency}, a point-wise confidence mask is generated for all radar points: $\mathbf{m}_{i}^{r}=\mathds{1}(|radial(\mathbf{v}_{gt,i})-\mathrm{ARV}_{i} | < \gamma_{thre})$. $\mathds{1}(\cdot)$ returns one when the condition in $(\cdot)$ is satisfied. Similarly, the filtered radar points are divided into three point sets: 
$\mathbf{S}_1^R \cup \mathbf{S}_2^R \cup \mathbf{S}_3^R =   \{\mathbf{p}_{i}|\mathbf{m}_{i}^{r}=1,\mathbf{p}_{i}\in\mathbf{P}^R\} $. Similar to Eq.~\eqref{eq:li_loss}, our masked radar flow loss is formulated as:
\begin{equation}
    \mathcal{L}_{ra} = \sum_{k=1}^{3}\frac{1}{|\mathbf{S}^{R}_{k}|} \sum_{p_{i} \in \mathbf{S}^{R}_{k}} \left\| \mathbf{V}^{R}(p_{i})- \mathbf{V}^{R}_{gt}(p_{i})\right\|_2.
\label{eq:ra_loss}
\end{equation}

\begin{table*}[ht!]
\centering
\resizebox{0.99\textwidth}{!}{
\begin{tabular}{l|c|c|ccccc|ccccc}
\hline
\multirow{3}{*}{Method} & \multirow{3}{*}{Sup.} & \multirow{3}{*}{Input} & \multicolumn{5}{c|}{\textbf{Lidar}} & \multicolumn{5}{c}{\textbf{Radar}} 
\\ \cline{4-13}  
&             &                               & \multicolumn{5}{c|}{Endpoint Error ($\downarrow$)} & \multicolumn{5}{c}{Endpoint Error ($\downarrow$)} 
\\ \cline{4-13}
&         &                      
& 3-way & 3D & FD & BS & FS & 3-way & 3D & FD & BS & FS

\\ \hline

FastFlow3D~\cite{jund2021scalable} &Full & L  
& $0.0789$ & $0.0491$ & $0.1625$   & $0.0348$ & $0.0395$ 
& -  & -  & -  & - & - 
\\
Flow4D~\cite{kim2025flow4d} &Full  & L 
& $0.0771$ & $0.0414$ & $0.1708$   & $0.0248$ & $0.0359$ 
& -  & -  & -  & - & - 
\\
Flow4D*~\cite{kim2025flow4d}  &Full & L  
& $0.0732$ & $0.0380$ & $0.1650$   & $0.0219$ & $0.0326$ 
& -  & -  & -  & - & -
\\
DeFlow~\cite{zhang2024deflow} &Full & L  
& $0.0691$ & $0.0321$ & $0.1562$   & $0.0152$ & $0.0359$  & -  & -  & -  & - & - 
\\ 
RaFlow~\cite{ding2022self} &Self  & R 
& -  & -  & -  & - & - 
& $0.1823$ & $0.1678$ & $0.2982$   & $0.1458$ & $0.1030$
\\
CMFlow~\cite{ding2023hidden}  & Cross & R 
& -  & -  & -  & - & - 
& $0.1180$ & $0.1242$ & $0.1652$   & $0.1187$ & $0.0699$
\\
\textbf{RaLiFlow} (Ours)  &Full  & L+R  
& $\textbf{0.0556}$ & $\textbf{0.0252}$ & $\textbf{0.1282}$   & $\textbf{0.0116}$ & $\textbf{0.0269}$ 
& $\textbf{0.0588}$ & $\textbf{0.0366}$ & $\textbf{0.1298}$   & $\textbf{0.0206}$ & $\textbf{0.0261}$  
\\
\hline
\end{tabular}}
\caption{Quantitative comparison on VoD validation set. RaFlow is self-supervised, and CMFlow requires cross-modal supervision from ego motion, dynamic segmentation, 2D optical flow and 3D scene flow labels; all the other methods are supervised by their input modalities. L denotes LiDAR input, and R denotes radar input.}
\label{tab:vod}
\end{table*}

\begin{figure*}[ht!]
\centering
\includegraphics[width=0.96\linewidth]{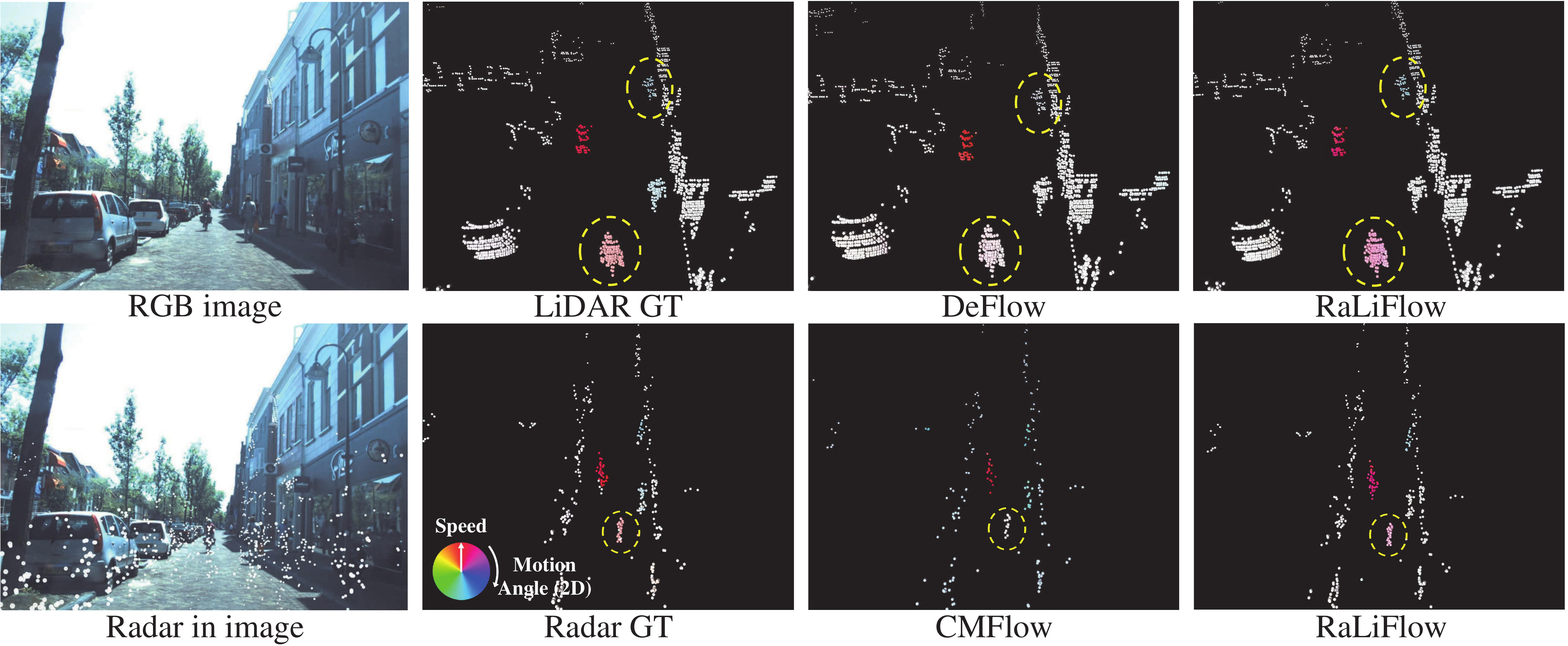}
\caption{Qualitative Results on VoD validation dataset. The top row and the bottom row display the LiDAR-based and radar-based scene flow estimation results.
The direction and magnitude of each flow vector are employed as hue and saturation, respectively. We enlarge the point size for better illustration, and the size of the radar point projected in the image reflects its depth in the 3D space.}
\label{fig:result}
\end{figure*}

\paragraph{Instance-level Dynamic Flow Consistency Loss.}
Scene flow estimation on featureless surfaces is challenging because the network tends to match the points in the source frame to nearby areas with similar features in the target frame, which may lead to underestimated flow predictions. 
To address this issue, we suggest that the flow estimation results within each foreground instance should be consistent with the maximum flow among them. 
Specifically, we first identity $H$ foreground instances $\{\mathbf{Ins}_{h}\}_{h=1}^{H}$ in the source frame based on the tracking boxes provided by the dataset. For each instance $\mathbf{Ins}_{h}$, we collect the corresponding radar points $\mathbf{P}_{h}^{R}$ and LiDAR points $\mathbf{P}_{h}^{L}$ located within it. Note that the valid dynamic radar points, as verified in Sec.~\ref{sf_gene}, are also assigned to their nearest foreground instance. We then select and merge the dynamic points from both modalities to form a dynamic point set: $\mathbf{P}_{h}^{D}=\{p_{c}~|\left\|\mathbf{V}_{gt}(p_{c})\right\|_2 \ge 0.5\mathrm{~m/s}, p_{c}\in(\mathbf{P}_{h}^{R} \cup \mathbf{P}_{h}^{L})\}$.  Next, we find the index of the point in $\mathbf{P}_{h}^{D}$ with the largest estimated scene flow:
\begin{equation}    
    \kappa = \arg\max_{c} \Bigl\{\left\|\mathbf{V}(p_{c})\right\|_2 ~~\Big| ~~ p_{c} \in \mathbf{P}_{h}^{D}\Bigr\}. \label{eq:k}
\end{equation}
Based on this, we define the instance-level dynamic flow consistency loss as:
\begin{equation}
    \mathcal{L}_{ins} = \sum_{h=1}^{H}\frac{1}{|\mathbf{P}_{h}^{D}|} \sum_{p_{c} \in \mathbf{P}_{h}^{D}} \left\| \mathbf{V}(p_{c})- \mathbf{V}(p_{\kappa})\right\|_2.
\label{eq:ins_loss}
\end{equation}
Finally, the total loss for RaLiFlow incorporates all three losses introduced above: 
\begin{equation}
\mathcal{L}_{total} = \mathcal{L}_{li}+\mathcal{L}_{ra}+\mathcal{L}_{ins}.
\end{equation}

\section{Experiments}
\label{sc:exp}

\subsection{Experimental Setup}
\noindent\textbf{Datasets.}
With the data prepocessing procedure proposed in Sec.~\ref{sc:data}, we derive a Radar-LiDAR-based scene flow dataset from the View-of-Delft (VoD) dataset~\cite{palffy2022multi}. Currently, the VoD dataset is the only public automotive dataset that contains synchronized and calibrated real-world data of LiDAR and 4D radar point clouds, ego vehicle’s odometry, images, and 3D tracking box annotations frame by frame (total 8,600 frames). We follow the official split of the VoD dataset and generate scene flow labels for the training and validation set. The unlabeled test set is unused in our case. Following previous works~\cite{ding2022self,ding2023hidden} on VoD dataset, only points within the viewing frustum are reserved, because the tracking annotation and radar data are not available in other regions.

\noindent\textbf{Implementation Details.}
\label{sc:implement}
For a fair comparison with the previous works, we follow ~\cite{zhang2024deflow} and use four GRU iterations for flow prediction. We set $1/{\sigma^2}=10$ for computation efficiency. With a resolution of $0.1\mathrm{~m}$ for each pillar encoder, we train our RaLiFlow and other pillar-based 
methods~\cite{jund2021scalable,zhang2024deflow} for 150 epochs.
To compare with voxel-based Flow4D~\cite{kim2025flow4d}, we implement both the official 2-frame version and a corresponding Flow4D* version, which sets the edge length of the voxel cube to $0.2\mathrm{~m}$ and $0.1\mathrm{~m}$ during voxelization, respectively. These voxel-based models are trained for 400 epochs. 
During training, we use Adam optimizer with an initial learning rate of $2\times 10^{-6}$. 
In addition, we follow ~\cite{ding2023hidden} to generate the extra required optical flow labels and train CMFlow on our proposed dataset under fully-supervised settings. Following official settings on VoD dataset, RaFlow~\cite{ding2022self} and CMFlow are trained for 50 and 75 epochs, respectively. All the methods are trained on the VoD training dataset.

\noindent\textbf{Evaluation Metrics.}
The End Point Error (EPE) measures the L2
norm of the discrepancy between the scene flow predictions and the ground truth flow vectors. Following previous works ~\cite{jund2021scalable,zhang2024deflow,zhang2024seflow,khoche2025ssf,kim2025flow4d}, we adopt the three-way Endpoint
Error (3-way EPE) for evaluation, which computes the unweighted average EPE of points located in Foreground Dynamic (`$\mathcal{FD}$'), Background Static (`$\mathcal{BS}$') and Foreground Static (`$\mathcal{FS}$') regions. For a convenient comparison with other existing methods which do not support 3-way EPE evalution, we also present the commonly used 3D end-point-error (3D EPE) results, which is the averaged EPE result over all input points in the source frame.

\subsection{Main Results}
\label{sc:quantitative}

\paragraph{Quantitative Results.} 
Since there is currently no other method that supports joint scene flow estimation on both radar and LiDAR point clouds, we compare our proposed method with other state-of-the-art single-modal methods in this section. Extensive comparisons are made with the pillar-based FastFlow3D~\cite{jund2021scalable} and DeFlow~\cite{zhang2024deflow}, the voxel-based Flow4D~\cite{kim2025flow4d} as well as the point-based RaFlow~\cite{ding2022self} and CMFlow~\cite{ding2023hidden}. The experimental results in Table~\ref{tab:vod} demonstrate that our RaLiFlow achieves the leading position on VoD dataset with its novel and powerful radar-LiDAR fusion strategy and loss design. Compared with the radar-only baseline CMFlow~\cite{ding2023hidden}, our RaLiFlow presents a significant performance improvement of 70.5\% on the 3D EPE metric and a large error reduction of 50.2\% on the 3-way EPE metric. Compared with LiDAR-only baselines, our RaLiFlow outperforms the best DeFlow~\cite{zhang2024deflow} by a remarkable margin of 21.5\% on the 3D EPE metric and 19.5\% on the 3-way EPE metric. Note that our RaLiFlow can dynamically adapt to large scale real-world data inputs; RaFlow~\cite{ding2022self} and CMFlow~\cite{ding2023hidden}, on the other hand, only supports a fixed input size of 256 points. The quantitative results in Table~\ref{tab:vod} demonstrate the effectiveness of our RaLiFlow architecture and loss design.

\paragraph{Qualitative Results.} 
In Fig.~\ref{fig:result}, we compare the qualitative outcomes of our RaLiFlow with scene flow groundtruth (GT) and the state-of-the-art single-modal methods. While DeFlow~\cite{zhang2024deflow} fails to estimate the dynamic flows on pedestrians circled in yellow, our method generates more sensible scene flow prediction results. It can also be observed that CMFlow~\cite{ding2023hidden} is limited by the fixed size of the input radar point cloud and produces less reliable and sparser flows. Specifically, CMFlow~\cite{ding2023hidden} wrongly estimates the flows in the background and produces inaccurate flows for the pedestrian circled in yellow. Compared with other methods, our RaLiflow generates scene flow predictions that best match the ground truth.

\begin{table*}[t!]
\centering
\resizebox{0.99\textwidth}{!}{
\begin{tabular}{l|ccccc|ccccc}
\hline
\multirow{3}{*}{Fusion Method} & \multicolumn{5}{c|}{\textbf{Lidar}} & \multicolumn{5}{c}{\textbf{Radar}} 
\\ \cline{2-11}  
& \multicolumn{5}{c|}{Endpoint Error ($\downarrow$)} & \multicolumn{5}{c}{Endpoint Error ($\downarrow$)} 
\\ \cline{2-11}
& 3-way & 3D & FD & BS & FS & 3-way & 3D & FD & BS & FS
\\ \hline
Concat
& $0.0603$ & $0.0294$ & $0.1358$   & $0.0153$ & $0.0297$ 
& $0.0634$ & $0.0408$ & $0.1371$   & $0.0242$ & $0.0289$ 
\\
DBCF w/o Gaussian 
& $0.0562$ & $0.0261$ & $0.1292$   & $0.0127$ & \textbf{0.0269}
& $0.0596$ & $0.0374$ & $0.1312$   & $0.0214$ & $0.0262$ 
\\
DBCF with Gaussian  
& $\textbf{0.0556}$ & $\textbf{0.0252}$ & $\textbf{0.1282}$   & $\textbf{0.0116}$ & $\textbf{0.0269}$ 
& $\textbf{0.0588}$ & $\textbf{0.0366}$ & $\textbf{0.1298}$   & $\textbf{0.0206}$ & $\textbf{0.0261}$  
\\
\hline
\end{tabular}}
\caption{Ablation study of Fusion methods on VoD validation set.}
\label{tab:ablation}
\end{table*}

\begin{table*}[h!]
\centering
\resizebox{0.99\textwidth}{!}{
\begin{tabular}{l|c|ccccc|ccccc}
\hline
\multirow{3}{*}{$\mathbf{L}_{ra}$} & \multirow{3}{*}{$\mathbf{L}_{ins}$}  & \multicolumn{5}{c|}{\textbf{Lidar}} & \multicolumn{5}{c}{\textbf{Radar}} 
\\ \cline{3-12}  
         &                               & \multicolumn{5}{c|}{Endpoint Error ($\downarrow$)} & \multicolumn{5}{c}{Endpoint Error ($\downarrow$)} 
\\ \cline{3-12}
       &                      
& 3-way & 3D & FD & BS & FS & 3-way & 3D & FD & BS & FS
\\ \hline
masked  &  
& $0.0577$ & $0.0272$ & $0.1314$   & $0.0132$ & $0.0286$ 
& $0.0610$ & $0.0401$ & $0.1300$   & $0.0243$ & $0.0287$  
\\
w/o mask &  $\checkmark$
& $0.0559$ & $0.0254$ & $0.1289$   & $0.0118$ & $0.0270$ 
& $0.0596$ & $0.0368$ & $0.1320$   & $0.0206$ & $0.0263$
\\
masked   & $\checkmark$ 
& $\textbf{0.0556}$ & $\textbf{0.0252}$ & $\textbf{0.1282}$   & $\textbf{0.0116}$ & $\textbf{0.0269}$ 
& $\textbf{0.0588}$ & $\textbf{0.0366}$ & $\textbf{0.1298}$   & $\textbf{0.0206}$ & $\textbf{0.0261}$  
\\
\hline
\end{tabular}}
\caption{Ablation study of Loss functions on VoD validation set. $\mathcal{L}_{ra}$ w/o mask simply sets $\mathbf{m}_{i}^{r}=1$ for all radar points $\mathbf{p}_{i}\in\mathbf{P}^R$.}
\label{tab:loss_ablation}
\end{table*}

\begin{table*}[ht!]
\centering
\centering
\resizebox{0.95\textwidth}{!}{
\begin{tabular}{c|ccccc|ccccc}
\hline
 \multirow{3}{*}{Input} & \multicolumn{5}{c|}{Lidar} & \multicolumn{5}{c}{Radar} 
\\ \cline{2-11}  
            & \multicolumn{5}{c|}{Endpoint Error ($\downarrow$)} & \multicolumn{5}{c}{Endpoint Error ($\downarrow$)} 
\\ \cline{2-11}            
& 3-way & 3D & FD & BS & FS & 3-way & 3D & FD & BS & FS
\\ \hline
L+L  & $0.0673$ & $0.0292$ & $0.1555$   & $0.0130$ & $0.0333$  & -  & -  & -  & - & - 
\\ 
 R+R  & -  & -  & -  & - & -
& $0.0861$ & $0.0561$ & $0.1926$   & $0.0325$ & $0.0330$   
\\ 
 L+R  
& \textbf{$0.0562$} & \textbf{$0.0261$} & \textbf{$0.1292$ }  & \textbf{$0.0127$} & \textbf{$0.0269$}
& \textbf{$0.0596$} & \textbf{$0.0374$} & \textbf{$0.1312$}   & \textbf{$0.0214$} & \textbf{$0.0262$ }   
\\
\hline
\end{tabular}}
\caption{Ablation study on different combination of input modalities on VoD validation set. `FD' represents Foreground Dynamic, `BS' represents Background Static, and `FS' represents Foreground Static.  L denotes LiDAR input, and R denotes radar input. RaLiFlow$^{\dag}$ is trained for 150 epochs with a batch size of 8 for this ablation. }
\label{tab:input}
\end{table*}

\subsection{Ablation Studies}

\paragraph{DBCF Module.} 
We examine the design of DBCF module by replacing it with different radar-LiDAR fusion methods.
As shown in Table~\ref{tab:ablation}, our DBCF module with bidirectional cross-attention operation outperforms the simple concatenation of radar and LiDAR 2D feature maps. Compared to simple concatenation, DBCF integrated with attentive Gaussian weights from radar dynamics achieves an overall performance improvement of 7.8\% on the LiDAR 3-way EPE metric and 7.3\% on the radar 3-way EPE metric. 
Furthermore, the addition of radar dynamics improves the prediction accuracy of the DBCF in the most challenging $\mathcal{FD}$ region as well as the $\mathcal{BS}$ area, which illustrates that our novel DBCF better utilizes input radar data to promote a more comprehensive understanding of dynamic environments and achieve effective cross-modal fusion.

\paragraph{Loss Terms.} 
The exprimental results in Table~\ref{tab:loss_ablation} demonstrate that our masked $\mathcal{L}_{ra}$ is able to reduce the adverse impact of unreliable radar points during training, enhancing the network performance in the difficult $\mathcal{FD}$ region for radar. Moreover, incorporating the $\mathcal{L}_{ins}$ constraint effectively reduces the scene flow estimation error on geometrically featureless moving surfaces, resulting in a significant performance improvement of 7.4\% on the LiDAR 3D EPE metric and 8.7\% on the radar 3D EPE metric.

\paragraph{Combination of Input Modalities.} 
\label{sec:ablation}
To further validate the effectiveness of our radar-LiDAR fusion method, we evaluate the radar-only and LiDAR-only performance under our proposed framework by duplicating one modality to preserve network structure. Since the generation of Gaussian heatmap requires radar information which is unavailable under LiDAR-only settings, we conduct this ablation study on RaLiFlow$^{\dag}$, which is another version of original RaLiflow without Gaussian heatmap integrated in the DBCF module. For LiDAR-only input ablation, $\mathcal{L}_{li}$ and $\mathcal{L}_{ins}$ are employed for supervion; and for radar-only ablation, $\mathcal{L}_{ra}$ and $\mathcal{L}_{ins}$ are adopted for network training. 

As shown in Table~\ref{tab:input}, the performance of our RaLiFlow$^{\dag}$ with LiDAR and radar input outperforms that of single-modal input on all evaluation metircs. Compared with LiDAR-only and radar-only cases, our radar-LiDAR fusion strategy improves the overall scene flow estimation performance on the 3-way EPE metric by 16.5\% and 30.8\%, respectively.
This analysis verifies the complementary role of each sensor and the effectiveness of our cross-modal fusion design. Moreover, it can also be observed that the single-modal scene flow estimation results on our RaLiFlow$^{\dag}$ still outperform the corresponding prior state-of-the-art methods, demonstrating the advanced performance of our framework.

\section{Conclusion }
In this paper, we propose the first joint scene flow learning framework for 4D radar and LiDAR point clouds, namely RaLiFlow. We explore the complementary information interaction between these two modalities and achieve effective radar-LiDAR fusion via our novel DBCF module. We also delve deeper into the inherent characteristics of 4D radar data and propose a novel data preprocessing strategy to generate more reliable radar scene flow labels based on the VoD dataset. The significant performance improvement of RaLiFlow over existing single-modal methods on the repurposed scene flow dataset demonstrates that our joint scene flow learning architecture effectively promotes the synergistic integration of 4D radar and LiDAR information. We hope our work highlights the promising potential of radar-LiDAR fusion for scene flow learning in the future.

\section*{Acknowledgements}
This work was conducted during Jingyun’s visit to the IMPL Lab at SUTD, funded by China Scholarship Council. 
This research is supported by The Key Research $\&$ Development Plan of Zhejiang Province under Grant No.2024C01010, 2024C01017, 2025C01039, and the Joint R$\&$D Program of the Yangtze River Delta Community of Sci-Tech Innovation with grant number 2024CSJGG01000.
It is also supported by the Agency for Science, Technology and Research (A*STAR) under its MTC Programmatic Funds (Grant No. M23L7b0021).


\begin{thebibliography}{62}
\providecommand{\natexlab}[1]{#1}

\bibitem[{Chen et~al.(2025)Chen, Zhang, Hao, and Zhou}]{chen2025ssf}
Chen, Y.; Zhang, M.; Hao, Q.; and Zhou, G. 2025.
\newblock SSF-PAN: Semantic Scene Flow-Based Perception for Autonomous Navigation in Traffic Scenarios.
\newblock \emph{arXiv preprint arXiv:2501.16754}.

\bibitem[{Cheng and Ko(2022)}]{cheng2022bi}
Cheng, W.; and Ko, J.~H. 2022.
\newblock Bi-pointflownet: Bidirectional learning for point cloud based scene flow estimation.
\newblock In \emph{European Conference on Computer Vision}, 108--124. Springer.

\bibitem[{Cheng and Ko(2023)}]{cheng2023multi}
Cheng, W.; and Ko, J.~H. 2023.
\newblock Multi-scale bidirectional recurrent network with hybrid correlation for point cloud based scene flow estimation.
\newblock In \emph{Proceedings of the IEEE/CVF International Conference on Computer Vision}, 10041--10050.

\bibitem[{Cho et~al.(2014)Cho, Van~Merri{\"e}nboer, Bahdanau, and Bengio}]{cho2014properties}
Cho, K.; Van~Merri{\"e}nboer, B.; Bahdanau, D.; and Bengio, Y. 2014.
\newblock On the properties of neural machine translation: Encoder-decoder approaches.
\newblock \emph{arXiv preprint arXiv:1409.1259}.

\bibitem[{Desai, Schumann, and Alsheakhali(2020)}]{desai2020review}
Desai, N.; Schumann, T.; and Alsheakhali, M. 2020.
\newblock A review of PointPillars architecture for object detection from point clouds.
\newblock In \emph{2020 IEEE International Conference on Consumer Electronics-Taiwan (ICCE-Taiwan)}, 1--2. IEEE.

\bibitem[{Ding et~al.(2024)Ding, Luo, Zhao, and Lu}]{ding2024milliflow}
Ding, F.; Luo, Z.; Zhao, P.; and Lu, C.~X. 2024.
\newblock milliflow: Scene flow estimation on mmwave radar point cloud for human motion sensing.
\newblock In \emph{European Conference on Computer Vision}, 202--221. Springer.

\bibitem[{Ding et~al.(2023)Ding, Palffy, Gavrila, and Lu}]{ding2023hidden}
Ding, F.; Palffy, A.; Gavrila, D.~M.; and Lu, C.~X. 2023.
\newblock Hidden gems: 4d radar scene flow learning using cross-modal supervision.
\newblock In \emph{Proceedings of the IEEE/CVF Conference on Computer Vision and Pattern Recognition}, 9340--9349.

\bibitem[{Ding et~al.(2022{\natexlab{a}})Ding, Pan, Deng, Deng, and Lu}]{ding2022self}
Ding, F.; Pan, Z.; Deng, Y.; Deng, J.; and Lu, C.~X. 2022{\natexlab{a}}.
\newblock Self-supervised scene flow estimation with 4-d automotive radar.
\newblock \emph{IEEE Robotics and Automation Letters}, 7(3): 8233--8240.

\bibitem[{Ding et~al.(2022{\natexlab{b}})Ding, Dong, Xu, Xu, Wang, and Li}]{ding2022fh}
Ding, L.; Dong, S.; Xu, T.; Xu, X.; Wang, J.; and Li, J. 2022{\natexlab{b}}.
\newblock Fh-net: A fast hierarchical network for scene flow estimation on real-world point clouds.
\newblock In \emph{European Conference on Computer Vision}, 213--229. Springer.

\bibitem[{Er{\c{c}}elik et~al.(2022)Er{\c{c}}elik, Yurtsever, Liu, Yang, Zhang, Top{\c{c}}am, Listl, Cayl{\i}, and Knoll}]{erccelik20223d}
Er{\c{c}}elik, E.; Yurtsever, E.; Liu, M.; Yang, Z.; Zhang, H.; Top{\c{c}}am, P.; Listl, M.; Cayl{\i}, Y.~K.; and Knoll, A. 2022.
\newblock 3d object detection with a self-supervised lidar scene flow backbone.
\newblock In \emph{European Conference on Computer Vision}, 247--265. Springer.

\bibitem[{Geiger et~al.(2013)Geiger, Lenz, Stiller, and Urtasun}]{geiger2013vision}
Geiger, A.; Lenz, P.; Stiller, C.; and Urtasun, R. 2013.
\newblock Vision meets robotics: The kitti dataset.
\newblock \emph{The International Journal of Robotics Research}, 32(11): 1231--1237.

\bibitem[{Guo et~al.(2024)Guo, An, Yang, Liu, and Liu}]{guo2024fsf}
Guo, E.; An, P.; Yang, Y.; Liu, Q.; and Liu, A.-A. 2024.
\newblock FSF-Net: Enhance 4D Occupancy Forecasting with Coarse BEV Scene Flow for Autonomous Driving.
\newblock \emph{arXiv preprint arXiv:2409.15841}.

\bibitem[{Han et~al.(2024)Han, Zhao, Chen, Ma, and Zhang}]{han2024dual}
Han, Y.; Zhao, N.; Chen, W.; Ma, K.~T.; and Zhang, H. 2024.
\newblock Dual-perspective knowledge enrichment for semi-supervised 3d object detection.
\newblock In \emph{Proceedings of the AAAI Conference on Artificial Intelligence}, volume~38, 2049--2057.

\bibitem[{Himmelsbach, Hundelshausen, and Wuensche(2010)}]{himmelsbach2010fast}
Himmelsbach, M.; Hundelshausen, F.~V.; and Wuensche, H.-J. 2010.
\newblock Fast segmentation of 3D point clouds for ground vehicles.
\newblock In \emph{2010 IEEE Intelligent Vehicles Symposium}, 560--565. IEEE.

\bibitem[{Jiang et~al.(2024)Jiang, Wang, Liu, Wang, Ma, Liu, Liang, Shan, and Du}]{jiang20243dsflabelling}
Jiang, C.; Wang, G.; Liu, J.; Wang, H.; Ma, Z.; Liu, Z.; Liang, Z.; Shan, Y.; and Du, D. 2024.
\newblock 3dsflabelling: Boosting 3d scene flow estimation by pseudo auto-labelling.
\newblock In \emph{Proceedings of the IEEE/CVF Conference on Computer Vision and Pattern Recognition}, 15173--15183.

\bibitem[{Jiao, Tran, and Shi(2021)}]{jiao2021effiscene}
Jiao, Y.; Tran, T.~D.; and Shi, G. 2021.
\newblock Effiscene: Efficient per-pixel rigidity inference for unsupervised joint learning of optical flow, depth, camera pose and motion segmentation.
\newblock In \emph{Proceedings of the IEEE/CVF Conference on Computer Vision and Pattern Recognition}, 5538--5547.

\bibitem[{Jin et~al.(2022)Jin, Lei, Akhtar, Li, and Hayat}]{jin2022deformation}
Jin, Z.; Lei, Y.; Akhtar, N.; Li, H.; and Hayat, M. 2022.
\newblock Deformation and correspondence aware unsupervised synthetic-to-real scene flow estimation for point clouds.
\newblock In \emph{Proceedings of the IEEE/CVF Conference on Computer Vision and Pattern Recognition}, 7233--7243.

\bibitem[{Jund et~al.(2021)Jund, Sweeney, Abdo, Chen, and Shlens}]{jund2021scalable}
Jund, P.; Sweeney, C.; Abdo, N.; Chen, Z.; and Shlens, J. 2021.
\newblock Scalable scene flow from point clouds in the real world.
\newblock \emph{IEEE Robotics and Automation Letters}, 7(2): 1589--1596.

\bibitem[{Khatri et~al.(2024)Khatri, Vedder, Peri, Ramanan, and Hays}]{khatri2024can}
Khatri, I.; Vedder, K.; Peri, N.; Ramanan, D.; and Hays, J. 2024.
\newblock I can’t believe it’s not scene flow!
\newblock In \emph{European Conference on Computer Vision}, 242--257. Springer.

\bibitem[{Khoche et~al.(2025)Khoche, Zhang, Sanchez, Asefaw, Mansouri, and Jensfelt}]{khoche2025ssf}
Khoche, A.; Zhang, Q.; Sanchez, L.~P.; Asefaw, A.; Mansouri, S.~S.; and Jensfelt, P. 2025.
\newblock SSF: Sparse Long-Range Scene Flow for Autonomous Driving.
\newblock \emph{arXiv preprint arXiv:2501.17821}.

\bibitem[{Kim et~al.(2025)Kim, Woo, Shin, Oh, and Im}]{kim2025flow4d}
Kim, J.; Woo, J.; Shin, U.; Oh, J.; and Im, S. 2025.
\newblock Flow4D: Leveraging 4D Voxel Network for LiDAR Scene Flow Estimation.
\newblock \emph{IEEE Robotics and Automation Letters}.

\bibitem[{Kingma and Ba(2014)}]{kingma2014adam}
Kingma, D.~P.; and Ba, J. 2014.
\newblock Adam: A method for stochastic optimization.
\newblock \emph{arXiv preprint arXiv:1412.6980}.

\bibitem[{Lang et~al.(2019)Lang, Vora, Caesar, Zhou, Yang, and Beijbom}]{lang2019pointpillars}
Lang, A.~H.; Vora, S.; Caesar, H.; Zhou, L.; Yang, J.; and Beijbom, O. 2019.
\newblock Pointpillars: Fast encoders for object detection from point clouds.
\newblock In \emph{Proceedings of the IEEE/CVF conference on computer vision and pattern recognition}, 12697--12705.

\bibitem[{Li et~al.(2022)Li, Zheng, Giancola, and Ghanem}]{li2022sctn}
Li, B.; Zheng, C.; Giancola, S.; and Ghanem, B. 2022.
\newblock Sctn: Sparse convolution-transformer network for scene flow estimation.
\newblock In \emph{Proceedings of the AAAI conference on artificial intelligence}, volume~36, 1254--1262.

\bibitem[{Li et~al.(2023{\natexlab{a}})Li, Yang, Chen, Yang, Jin, and Akiyama}]{li2023pillardan}
Li, J.; Yang, L.; Chen, Y.; Yang, Y.; Jin, Y.; and Akiyama, K. 2023{\natexlab{a}}.
\newblock PillarDAN: Pillar-based Dual Attention Attention Network for 3D Object Detection with 4D RaDAR.
\newblock In \emph{2023 IEEE 26th International Conference on Intelligent Transportation Systems (ITSC)}, 1851--1857. IEEE.

\bibitem[{Li and Zhao(2024)}]{li2024end}
Li, L.; and Zhao, N. 2024.
\newblock End-to-end semi-supervised 3d instance segmentation with pcteacher.
\newblock In \emph{2024 IEEE International Conference on Robotics and Automation (ICRA)}, 5352--5358. IEEE.

\bibitem[{Li, Kaesemodel~Pontes, and Lucey(2021)}]{li2021neural}
Li, X.; Kaesemodel~Pontes, J.; and Lucey, S. 2021.
\newblock {Neural scene flow prior}.
\newblock \emph{NIPS}, 34: 7838--7851.

\bibitem[{Li et~al.(2023{\natexlab{b}})Li, Zheng, Ferroni, Pontes, and Lucey}]{li2023fast}
Li, X.; Zheng, J.; Ferroni, F.; Pontes, J.~K.; and Lucey, S. 2023{\natexlab{b}}.
\newblock Fast neural scene flow.
\newblock In \emph{Proceedings of the IEEE/CVF International Conference on Computer Vision}, 9878--9890.

\bibitem[{Liang et~al.(2025)Liang, Badki, Su, Tompkin, and Gallo}]{liang2025zero}
Liang, Y.; Badki, A.; Su, H.; Tompkin, J.; and Gallo, O. 2025.
\newblock Zero-Shot Monocular Scene Flow Estimation in the Wild.
\newblock \emph{arXiv preprint arXiv:2501.10357}.

\bibitem[{Lin et~al.(2024{\natexlab{a}})Lin, Xu, Wang, Wang, and Yang}]{lin2024flowmamba}
Lin, M.; Xu, G.; Wang, Y.; Wang, X.; and Yang, X. 2024{\natexlab{a}}.
\newblock FlowMamba: Learning Point Cloud Scene Flow with Global Motion Propagation.
\newblock \emph{arXiv preprint arXiv:2412.17366}.

\bibitem[{Lin and Caesar(2024)}]{lin2024icp}
Lin, Y.; and Caesar, H. 2024.
\newblock Icp-flow: Lidar scene flow estimation with icp.
\newblock In \emph{Proceedings of the IEEE/CVF Conference on Computer Vision and Pattern Recognition}, 15501--15511.

\bibitem[{Lin et~al.(2024{\natexlab{b}})Lin, Liu, Xia, Wang, Wang, Qi, Dong, Dong, Zhang, and Zhu}]{lin2024rcbevdet}
Lin, Z.; Liu, Z.; Xia, Z.; Wang, X.; Wang, Y.; Qi, S.; Dong, Y.; Dong, N.; Zhang, L.; and Zhu, C. 2024{\natexlab{b}}.
\newblock Rcbevdet: Radar-camera fusion in bird's eye view for 3d object detection.
\newblock In \emph{Proceedings of the IEEE/CVF Conference on Computer Vision and Pattern Recognition}, 14928--14937.

\bibitem[{Liu et~al.(2022)Liu, Lu, Xu, Liu, Li, and Chen}]{liu2022camliflow}
Liu, H.; Lu, T.; Xu, Y.; Liu, J.; Li, W.; and Chen, L. 2022.
\newblock Camliflow: bidirectional camera-lidar fusion for joint optical flow and scene flow estimation.
\newblock In \emph{Proceedings of the IEEE/CVF conference on computer vision and pattern recognition}, 5791--5801.

\bibitem[{Liu, Qi, and Guibas(2019)}]{liu2019flownet3d}
Liu, X.; Qi, C.~R.; and Guibas, L.~J. 2019.
\newblock Flow{N}et3{D}: {L}earning scene flow in 3{D} point clouds.
\newblock In \emph{CVPR}, 529--537.

\bibitem[{Liu, Shao, and Hoffmann(2021)}]{liu2021global}
Liu, Y.; Shao, Z.; and Hoffmann, N. 2021.
\newblock Global attention mechanism: Retain information to enhance channel-spatial interactions.
\newblock \emph{arXiv preprint arXiv:2112.05561}.

\bibitem[{Luo et~al.(2025)Luo, Cheng, Tang, Zhang, Xue, and Fan}]{luo2025mambaflow}
Luo, J.; Cheng, J.; Tang, X.; Zhang, Q.; Xue, B.; and Fan, R. 2025.
\newblock MambaFlow: A Novel and Flow-guided State Space Model for Scene Flow Estimation.
\newblock \emph{arXiv preprint arXiv:2502.16907}.

\bibitem[{Mayer et~al.(2016)Mayer, Ilg, Hausser, Fischer, Cremers, Dosovitskiy, and Brox}]{mayer2016large}
Mayer, N.; Ilg, E.; Hausser, P.; Fischer, P.; Cremers, D.; Dosovitskiy, A.; and Brox, T. 2016.
\newblock A large dataset to train convolutional networks for disparity, optical flow, and scene flow estimation.
\newblock In \emph{CVPR}, 4040--4048.

\bibitem[{Najibi et~al.(2022)Najibi, Ji, Zhou, Qi, Yan, Ettinger, and Anguelov}]{najibi2022motion}
Najibi, M.; Ji, J.; Zhou, Y.; Qi, C.~R.; Yan, X.; Ettinger, S.; and Anguelov, D. 2022.
\newblock Motion Inspired Unsupervised Perception and Prediction in Autonomous Driving.
\newblock In \emph{Computer Vision--ECCV 2022: 17th European Conference, Tel Aviv, Israel, October 23--27, 2022, Proceedings, Part XXXVIII}, 424--443. Springer.

\bibitem[{Palffy et~al.(2022)Palffy, Pool, Baratam, Kooij, and Gavrila}]{palffy2022multi}
Palffy, A.; Pool, E.; Baratam, S.; Kooij, J.~F.; and Gavrila, D.~M. 2022.
\newblock {Multi-class Road User Detection with 3+ 1D Radar in the View-of-Delft Dataset}.
\newblock \emph{RA-L}, 7(2): 4961--4968.

\bibitem[{Pan et~al.(2025)Pan, Cui, Yang, and Zhao}]{pan2025images}
Pan, Y.; Cui, Q.; Yang, X.; and Zhao, N. 2025.
\newblock How Do Images Align and Complement LiDAR? Towards a Harmonized Multi-modal 3D Panoptic Segmentation.
\newblock In \emph{Forty-second International Conference on Machine Learning}.

\bibitem[{Qi et~al.(2017{\natexlab{a}})Qi, Su, Mo, and Guibas}]{qi2017pointnet}
Qi, C.~R.; Su, H.; Mo, K.; and Guibas, L.~J. 2017{\natexlab{a}}.
\newblock Pointnet: Deep learning on point sets for 3d classification and segmentation.
\newblock In \emph{CVPR}, 652--660.

\bibitem[{Qi et~al.(2017{\natexlab{b}})Qi, Yi, Su, and Guibas}]{qi2017pointnet++}
Qi, C.~R.; Yi, L.; Su, H.; and Guibas, L.~J. 2017{\natexlab{b}}.
\newblock Pointnet++: Deep hierarchical feature learning on point sets in a metric space.
\newblock \emph{NIPS}, 30.

\bibitem[{Sheng et~al.(2022)Sheng, Cai, Zhao, Deng, Huang, Hua, Zhao, and Lee}]{sheng2022rethinking}
Sheng, H.; Cai, S.; Zhao, N.; Deng, B.; Huang, J.; Hua, X.-S.; Zhao, M.-J.; and Lee, G.~H. 2022.
\newblock Rethinking IoU-based optimization for single-stage 3D object detection.
\newblock In \emph{European Conference on Computer Vision}, 544--561. Springer.

\bibitem[{Sheng et~al.(2025)Sheng, Cai, Zhao, Deng, Liang, Zhao, and Ye}]{sheng2025ct3d++}
Sheng, H.; Cai, S.; Zhao, N.; Deng, B.; Liang, Q.; Zhao, M.-J.; and Ye, J. 2025.
\newblock CT3D++: Improving 3D Object Detection with Keypoint-Induced Channel-wise Transformer.
\newblock \emph{International Journal of Computer Vision}, 133(7): 4817--4836.

\bibitem[{Su et~al.(2018)Su, Jampani, Sun, Maji, Kalogerakis, Yang, and Kautz}]{su2018splatnet}
Su, H.; Jampani, V.; Sun, D.; Maji, S.; Kalogerakis, E.; Yang, M.-H.; and Kautz, J. 2018.
\newblock Splatnet: Sparse lattice networks for point cloud processing.
\newblock In \emph{CVPR}, 2530--2539.

\bibitem[{Tai et~al.(2024)Tai, Qian, Kang, Liu, and Liu}]{tai2024fusing}
Tai, H.; Qian, Y.; Kang, X.; Liu, L.; and Liu, Y. 2024.
\newblock Fusing LiDAR and Radar with Pillars Attention for 3D Object Detection.
\newblock In \emph{2024 7th International Symposium on Autonomous Systems (ISAS)}, 1--6. IEEE.

\bibitem[{Vedder et~al.(2023)Vedder, Peri, Chodosh, Khatri, Eaton, Jayaraman, Liu, Ramanan, and Hays}]{vedder2023zeroflow}
Vedder, K.; Peri, N.; Chodosh, N.; Khatri, I.; Eaton, E.; Jayaraman, D.; Liu, Y.; Ramanan, D.; and Hays, J. 2023.
\newblock ZeroFlow: scalable scene flow via distillation.
\newblock \emph{arXiv preprint arXiv:2305.10424}.

\bibitem[{Vedder et~al.(2024)Vedder, Peri, Khatri, Li, Eaton, Kocamaz, Wang, Yu, Ramanan, and Pehserl}]{vedder2024scene}
Vedder, K.; Peri, N.; Khatri, I.; Li, S.; Eaton, E.; Kocamaz, M.~K.; Wang, Y.; Yu, Z.; Ramanan, D.; and Pehserl, J. 2024.
\newblock Scene Flow as a Partial Differential Equation.
\newblock In \emph{The Thirteenth International Conference on Learning Representations}.

\bibitem[{Wan et~al.(2023)Wan, Mao, Zhang, and Dai}]{wan2023rpeflow}
Wan, Z.; Mao, Y.; Zhang, J.; and Dai, Y. 2023.
\newblock Rpeflow: Multimodal fusion of rgb-pointcloud-event for joint optical flow and scene flow estimation.
\newblock In \emph{Proceedings of the IEEE/CVF International Conference on Computer Vision}, 10030--10040.

\bibitem[{Wang et~al.(2022)Wang, Zhang, Xv, Zhang, Fu, Wang, Zhu, Ren, Lu, Li et~al.}]{wang2022interfusion}
Wang, L.; Zhang, X.; Xv, B.; Zhang, J.; Fu, R.; Wang, X.; Zhu, L.; Ren, H.; Lu, P.; Li, J.; et~al. 2022.
\newblock InterFusion: Interaction-based 4D radar and LiDAR fusion for 3D object detection.
\newblock In \emph{2022 IEEE/RSJ International Conference on Intelligent Robots and Systems (IROS)}, 12247--12253. IEEE.

\bibitem[{Wang et~al.(2025)Wang, Zhao, Han, Guo, and Yang}]{wang2025augrefer}
Wang, X.; Zhao, N.; Han, Z.; Guo, D.; and Yang, X. 2025.
\newblock Augrefer: Advancing 3d visual grounding via cross-modal augmentation and spatial relation-based referring.
\newblock In \emph{Proceedings of the AAAI Conference on Artificial Intelligence}, volume~39, 8006--8014.

\bibitem[{Wang et~al.(2023{\natexlab{a}})Wang, Chi, Lin, and Yang}]{wang2023ihnet}
Wang, Y.; Chi, C.; Lin, M.; and Yang, X. 2023{\natexlab{a}}.
\newblock Ihnet: Iterative hierarchical network guided by high-resolution estimated information for scene flow estimation.
\newblock In \emph{Proceedings of the IEEE/CVF International Conference on Computer Vision}, 10073--10082.

\bibitem[{Wang et~al.(2023{\natexlab{b}})Wang, Deng, Li, Hu, Liu, Zhang, Ji, Ouyang, and Zhang}]{wang2023bi}
Wang, Y.; Deng, J.; Li, Y.; Hu, J.; Liu, C.; Zhang, Y.; Ji, J.; Ouyang, W.; and Zhang, Y. 2023{\natexlab{b}}.
\newblock Bi-lrfusion: Bi-directional lidar-radar fusion for 3d dynamic object detection.
\newblock In \emph{Proceedings of the IEEE/CVF Conference on Computer Vision and Pattern Recognition}, 13394--13403.

\bibitem[{Wilson et~al.(2023)Wilson, Qi, Agarwal, Lambert, Singh, Khandelwal, Pan, Kumar, Hartnett, Pontes et~al.}]{wilson2023argoverse}
Wilson, B.; Qi, W.; Agarwal, T.; Lambert, J.; Singh, J.; Khandelwal, S.; Pan, B.; Kumar, R.; Hartnett, A.; Pontes, J.~K.; et~al. 2023.
\newblock Argoverse 2: Next generation datasets for self-driving perception and forecasting.
\newblock \emph{arXiv preprint arXiv:2301.00493}.

\bibitem[{Wu, Qi, and Fuxin(2019)}]{wu2019pointconv}
Wu, W.; Qi, Z.; and Fuxin, L. 2019.
\newblock Pointconv: Deep convolutional networks on 3d point clouds.
\newblock In \emph{CVPR}, 9621--9630.

\bibitem[{Wu et~al.(2020)Wu, Wang, Li, Liu, and Fuxin}]{wu2020pointpwc}
Wu, W.; Wang, Z.~Y.; Li, Z.; Liu, W.; and Fuxin, L. 2020.
\newblock Pointpwc-net: Cost volume on point clouds for (self-) supervised scene flow estimation.
\newblock In \emph{Computer Vision--ECCV 2020: 16th European Conference, Glasgow, UK, August 23--28, 2020, Proceedings, Part V 16}, 88--107. Springer.

\bibitem[{Xu et~al.(2021)Xu, Zhang, Wang, Hu, Li, Pan, Li, and Deng}]{xu2021rpfa}
Xu, B.; Zhang, X.; Wang, L.; Hu, X.; Li, Z.; Pan, S.; Li, J.; and Deng, Y. 2021.
\newblock RPFA-Net: A 4D radar pillar feature attention network for 3D object detection.
\newblock In \emph{2021 IEEE International Intelligent Transportation Systems Conference (ITSC)}, 3061--3066. IEEE.

\bibitem[{Xu et~al.(2024)Xu, Xiang, Zhang, Zhong, Zhao, Dang, Xu, Pu, and Liu}]{xu2024sckd}
Xu, R.; Xiang, Z.; Zhang, C.; Zhong, H.; Zhao, X.; Dang, R.; Xu, P.; Pu, T.; and Liu, E. 2024.
\newblock SCKD: Semi-Supervised Cross-Modality Knowledge Distillation for 4D Radar Object Detection.
\newblock \emph{arXiv preprint arXiv:2412.14571}.

\bibitem[{Zhang et~al.(2024{\natexlab{a}})Zhang, Yang, Fang, Geng, and Jensfelt}]{zhang2024deflow}
Zhang, Q.; Yang, Y.; Fang, H.; Geng, R.; and Jensfelt, P. 2024{\natexlab{a}}.
\newblock DeFlow: decoder of scene flow network in autonomous driving.
\newblock In \emph{2024 IEEE International Conference on Robotics and Automation (ICRA)}, 2105--2111. IEEE.

\bibitem[{Zhang et~al.(2024{\natexlab{b}})Zhang, Yang, Li, Andersson, and Jensfelt}]{zhang2024seflow}
Zhang, Q.; Yang, Y.; Li, P.; Andersson, O.; and Jensfelt, P. 2024{\natexlab{b}}.
\newblock Seflow: A self-supervised scene flow method in autonomous driving.
\newblock In \emph{European Conference on Computer Vision}, 353--369. Springer.

\bibitem[{Zhang et~al.(2023)Zhang, Zhang, Yu, Yi, Xie, and Ma}]{zhang2023lidar}
Zhang, Z.; Zhang, Z.; Yu, Q.; Yi, R.; Xie, Y.; and Ma, L. 2023.
\newblock Lidar-camera panoptic segmentation via geometry-consistent and semantic-aware alignment.
\newblock In \emph{Proceedings of the IEEE/CVF International Conference on Computer Vision}, 3662--3671.

\bibitem[{Zhao, Chua, and Lee(2021)}]{zhao2021few}
Zhao, N.; Chua, T.-S.; and Lee, G.~H. 2021.
\newblock Few-shot 3d point cloud semantic segmentation.
\newblock In \emph{Proceedings of the IEEE/CVF conference on computer vision and pattern recognition}, 8873--8882.

\end{thebibliography}
\end{document}